\documentclass[conference]{IEEEtran}
\usepackage{blindtext, graphicx}
\usepackage{caption}
\usepackage{subcaption}

\usepackage{url}
\begin{document}
%
\title{Parking Stall Vacancy Indicator System Based on Deep Convolutional Neural Networks}

\author{\IEEEauthorblockN{Sepehr Valipour, Mennatullah Siam, Eleni Stroulia, Martin Jagersand}
\IEEEauthorblockA{School of Computing Science\\
University of Alberta\\
\{valipour,mennatul\}@ualberta.ca}
}


\maketitle

\begin{abstract}
Parking management systems, and vacancy-indication services in particular, can play a valuable role in reducing traffic and energy waste in large cities. Visual detection methods represent a cost-effective option, since they can take advantage of hardware usually already available in many parking lots, namely cameras. However, visual detection methods can be fragile and not easily generalizable. In this paper, we present a robust detection algorithm based on deep convolutional neural networks. We implemented and tested our algorithm on a large baseline dataset, and also on a set of image feeds from actual cameras already installed in parking lots. We have developed a fully functional system, from server-side image analysis to front-end user interface, to demonstrate the practicality of our method.

\end{abstract}

\begin{IEEEkeywords}
Smart Cities, Smart Parking, Deep Learning, Internet of Things 
\end{IEEEkeywords}

%
\IEEEpeerreviewmaketitle

\section{Introduction}

Nowadays, smart devices have found their place in many aspects of our daily routines. Sensors analyze air particles to monitor levels of pollutants; embedded devices in cars control speed, recognize obstacles and help with maneuvering; smart watches monitor physiological parameters and control our ambient environment. This trend will only be growing as constant improvements in hardware, both in computation power and price, make these devices even more ubiquitous. 

Urban areas can greatly benefit from this new trend in technology \cite{zanella2014internet}\cite{pala2007smart}. 
%
%
%
%
%
Some of the fields that this paradigm can be applied to are transportation, lighting, surveillance and city planning. Among these, parking management with smart devices is gaining popularity \cite{lu2009spark}\cite{geng2013new}. Finding empty parking slot has become an everyday chore for many drivers in large cities. Traditional method of circling around the parking lots or streets to find a spot (Blind Search \cite{wang2011reservation}) is inefficient, time consuming and frustrating. Based on an study \cite{arnott2006integrated} nearly 30$\%$ of  traffic in cities is from cars that are cruising for parking, which, on average, lasts 7.8 minutes for each car. Parking vacancy indicators and guidance systems have multiple benefits. As a direct impact, it reduces time consumption and frustration for the driver, and as secondary effect, it alleviates overall traffic in cities and therefore reduces total fuel consumption and CO emission.

Different approaches in Parking Guidance and Information (PGI) systems vary greatly. The ultimate purpose of all PGI systems is to collect the most reliable and accurate data from parking vacancies and present them in such a way that it would be most useful for users. What set them apart are their broadcasting method, detection and optimization goals. However, the right vehicle detector system may be the most important part of a successful PGI. 

Three main parameters are to be considered when choosing a detection approach. First is reliability. A reliable detector should correctly report the status of the parking slot that it is monitoring regardless of changes in environmental parameters such as temperature, different vehicles and location. Second is the Installation cost that is attributed to cost of new equipment to be purchased per stall, difficulties in deployment and the need for limiting regular parking operation during the installment. Finally, cost of maintenance is the third factor. The number of sensing units has a direct effect on Installation and maintenance costs. A common way for detection is to use sensors such as ultrasonic, inductive loops and infrared lasers sensor, where all of which need to be installed per-stall. These sensors are usually reliable however, due to large number of parking stalls in parking lots, even minute cost for installation or maintenance per sensor can sum up to to a large amount.
%
%
%
%
%

\begin{figure}[ht]
    \centering
    \begin{subfigure}[b]{0.33\textwidth}
        \centering
        \includegraphics[width=\textwidth]{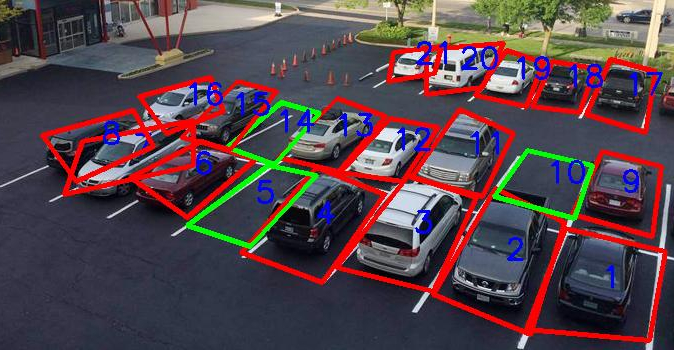}
    \end{subfigure} \begin{subfigure}[b]{0.13\textwidth}
        \centering
        \includegraphics[width=\textwidth]{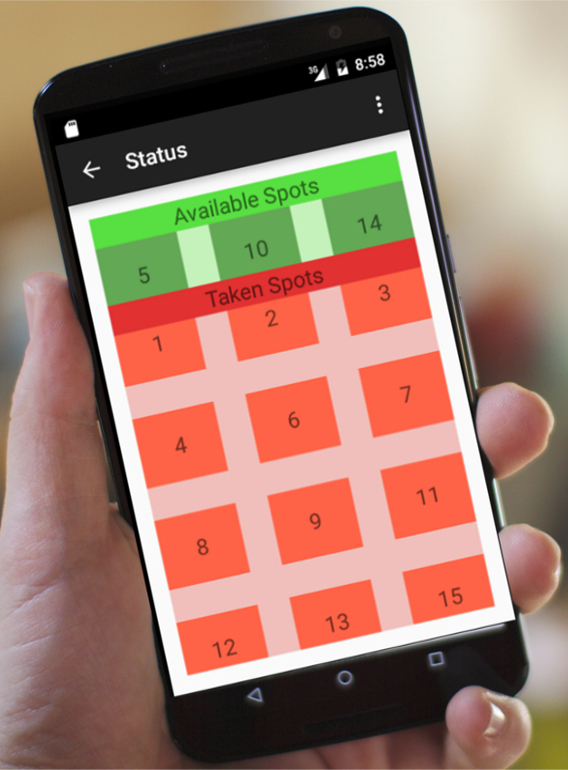}
    \end{subfigure}
    \caption{Free parking spaces, and proportion of total spaces are shown by our mobile app}
%
%
%
%
%
    \label{fig:intro}
\end{figure}

Considering drawbacks of sensor per-stall detection approach, vision-based detectors are relatively cost efficient. The installation is simple and requires no shutdowns. Each visual node, consisting of a camera and a transmitter, can monitor many vehicles simultaneously, lowering the cost per stall. Since they can be used for other purposes such as surveillance, the infrastructure is often already installed or it can be used by other applications after deployment. Since visual nodes have no physically engaged element, they require almost no maintenance. Even with all these advantages, reliability of visual systems have crippled their wide usage in industry. Many parameters can affect visual system detection such as light intensity, camera resolution and bad weather. As a result, most of the current research in this area is devoted to improve the robustness of visual PGI systems.

In this paper we are proposing a visual parking vacancy indicator system that utilizes deep Convolutional Neural Networks (CNN) for detection of vacant and occupied parking slots. We show the robustness of our detection system by testing it on a large dataset of labeled parking spots. To evaluate the practicality of this approach we developed the whole system from detection module to the front-end. Figure \ref{fig:intro} illustrates how the finished application works. In what follows, we first cover some related work and background. Then the architecture of the system is explained followed by experimental results. We finish by conclusion and future work.  

\section{Background and Related Work}
\subsection{Parking Guidance and Information}
Efforts for improving the PGI system can be divided into three groups. Detection, presentation and optimization.  

Many different sensors were used for parking vacancy detection \cite{mimbela2000summary}\cite{idris2009саг}. For single stall detection, ultrasonic sensors\cite{kianpisheh2012smart}, inductive loop detectors \cite{idris2009саг} and, more recently, in pavement wireless sensor networks \cite{bajwa2011pavement} are popular options. Due to accessibility of cameras and processing units for them, many attempts were made to use visual feedback for vacancy detection \cite{lin2006vision} \cite{ichihashi2010improvement} \cite{huang2010hierarchical}. A very recent work collected a large dataset of images from parking lots and applied a learning algorithm for vacancy detection with with acceptable results \cite{de2015pklot}. 

Variable Message Signs is a common practice for presenting space availability to drivers. These signs are usually placed on intersections and streets to inform the drivers about vacancies at nearby parking lots. An important drawback of this method is its limitation in conveying data which usually is restricted to number of vacant spots. More recent presentation methods leverage Internet to publish their data. Street Line \cite{streetline} is an industrialized example of such presentation method. 

%
%
%
%
%
Optimization methods provide the driver with a parking spot based on pre-defined or user-defined objectives such as proximity to the spot \cite{geng2013new}.

\subsection{Convolutional Neural Networks}
Traditionally, a combination of hand crafted features such as SIFT, ORB and BRISK \cite{nixon2008feature} plus a classifier, commonly SVM and Random Forrest, is used for the detection and recognition task. Accordingly, most of the state of the art research in visual PGI are using these methods. However, recent advances in computer vision in past few years, specifically deep learning, have improved traditional state of the art by a large margin for many visual tasks. Object recognition in particular has improved, and accurate methods are now available for it\cite{simonyan2014very} \cite{krizhevsky2012imagenet}.

\begin{figure}[ht]
    \centering
    \includegraphics[width=0.45 \textwidth]{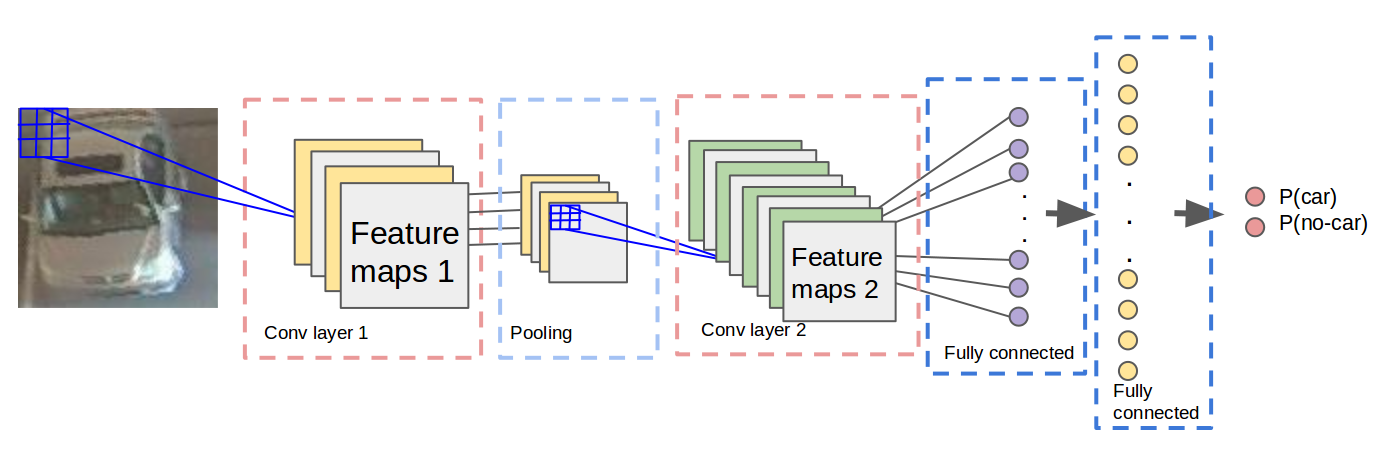}
    \caption{Overview of a deep neural network consisted of convolutional layers, activation layer, pooling layer and fully connected layers.}
    \label{fig:cnnarchitecture}
\end{figure}

Convolutional Neural Networks (CNN) can be recognized as an extension to regular Artificial Neural Networks (ANN) \cite{yegnanarayana2009artificial}. The main difference between these two methods is the usage of convolutional layers and pooling layers in CNN. In the convolutional layers, the value of each hidden unit is not just a linear transformation of all the hidden units of previous layer, which is the case for fully connected layers in ANN. Instead, the value is a result of convolving a three dimensional filter with values of previous layer. The Pooling layer is a maximum spatial response filter that passes the maximum values of region in the input layer to the output. See Figure \ref{fig:cnnarchitecture}.

These two innovations let CNNs have more trainable layers compare to ANN, hence the name deep. Learned filters in convolutional layers are convolved with the entire feature map. Therefore, size of these filters are not commensurate to the spatial size of feature maps, as it was the case for ANN. Accordingly, it dramatically shrinks search space for each layer. Pooling layers effectively reduce the spatial size of its input by the assumption that spatially close features are co-related, which is mostly true for images, and therefore one of them can represent them all. It also makes the network less sensitive to translation of the input image. Based on these fundamental elements, many different networks have been designed and trained on large datasets for image recognition tasks. A few of the most successful ones are GoogleNet \cite{szegedy2015going}, VGGNet \cite{chatfield2014return} and AlexNet \cite{krizhevsky2012imagenet}.
%
%
%
%
%

\section{Proposed Architecture}
\begin{figure}[ht]
    \centering
    \includegraphics[width=0.45 \textwidth]{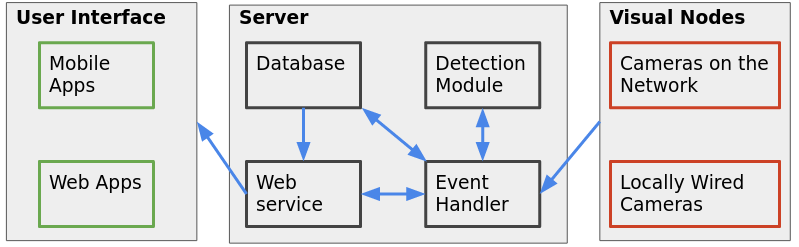}
    \caption{System architecture. The server collects data from visual nodes, then feed them to the detection module and updates the database using this information. It serves web service that front-end applications use to access the database.}
    \label{fig:architecture}
\end{figure}

Our proposed system is composed of three parts. First is the visual nodes, namely cameras, that are connected to the server through either local wireless network or through the Internet. The server is the second component. It includes a database, detection module, web service and event handler. The server collects images from visual nodes, feeds them to the detection module and collects and stores its output in the database and provides web service for front-end applications to get information from the database. The third element is the front-end that presents parking lots vacancies to users. Figure \ref{fig:architecture} visualizes the systems architecture. In the following each element is discussed in detail.

\subsection{Visual Nodes}
Regular color camera images are used in this project while there exist other choices such as microwave radar. As discussed, camera is a suitable choice for parking management system, due to its low maintenance, low cost per stall and ease of scalability. However, a drawback is that camera raw output images are highly sensitive to environmental parameters. Ranging from different light (time of the day) to different weather. Adding to these, we should mention camera parameters and point of view. Camera parameters include, camera intrinsic parameters, image size, imaging frequency and low level filtering such as noise cancellation. These parameters vary drastically between each camera model and manufacturer. Camera point of view is also effective in terms of what view of the stall is being observed.\\
We realized that restricting the cameras to certain specification is not practical. Parking owners in different locations will decide on the camera types that they want to use based on their budget and local availability. Cameras point of view is also dependent on the parking's structure and contractors who install the cameras. \\
Therefore, the only constraints that we put on our visual nodes are:
\begin{enumerate}
    \item Stalls of interest should not be visually blocked.
    \item Cameras image output should be  delivered upon server's request. 
\end{enumerate}
These expectations are easily satisfiable in practice. Visibility of stalls is common sense and it is generally taken into account by most surveillance systems. Second constraint, is a standard in almost all digital cameras where an embedded flash memory holds, at least, last taken image to provide on request. In our experiments, we used publicly available IP cameras in parking lots from different countries where we had no hand in installation or choosing the location or camera model.\\

\subsection{Server}
The server in our system has four responsibilities:
First, it is to host the database. A relational database is used in this system. It stores separate tables for different parking lot where each element in the table corresponds to a stall. Each stall has four mandatory fields. 1) Stall ID. A unique number in the parking lot. 2) Stalls bounding box coordinates (in image space). Coordinates are entered through our GUI by the administrator. 3) Image blob. Cropped image of the stall from current visual feed. This will be updated by the server on fixed intervals 4) Status. A binary value indicating vacant or occupied. This is also being updated by the server from the detection method's response.\\
Second, it is collecting data from cameras. Cameras can be connected to the server in a local network or can be connected through Internet. If cameras do not use HTTP protocol, a local communication protocol will be used on the server. If they do, requests and responses can be done with HTTP. If the cameras are connected to Internet, there is no need for the server to be located close to visual node and a server on the cloud is a valid option. \\
Third, it is serving a web-service. It bridges the database to our system's front-end. The main functionality supported is retrieving status of all stalls in each parking lot.\\
Finally, it feeds the images from visual nodes to the detection module along with bounding boxes of stalls and receives the detection module predictions.

\subsection{Detection Module}
The detection module is responsible for reporting the occupancy status of a parking stall given the image of the stall. We use a Convolutional Neural Network for this task.\\ 
Having few constraints on the input data puts a heavy burden on the detection system. Therefore robustness and generality of the detection algorithm have the highest priority. To achieve this, choosing the right network and training procedure is crucial. The design of our network is based on VGGNet-F \cite{chatfield2014return}. It has five convolutional layers where each is followed by a pooling layer and Rectified Linear activation function. It has three fully connected layers at the end that use the features from the convolutional layers for classification. The VGGNet architecture features a simple and uniform design throughout the network. Filters' kernel sizes are all 3 beside first two that are 11 and 5 respectively to reduce the network's size. VGGNet-F, which is the smallest of the VGG networks family, was chosen. Even though it is relatively small it was able to achieve 16.7$\%$ top five error on ILSVRC-2012 dataset \cite{ILSVRC15}.\\
This network is originally designed for classifying 1000 objects. We modified the last fully connected layer to output binary values for occupied or empty stall. This will reduce number of parameters of the network and decrease its effective size on the drive to 86MB. The detection module can either be placed inside the server or it can run on the cloud and only communicate with the server. 

\subsection{User Interface}
\begin{figure}[ht]
    \centering
    \includegraphics[width=0.45 \textwidth]{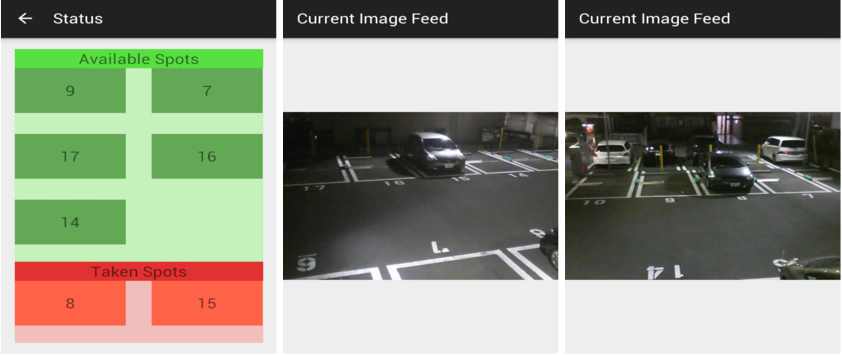}
    \caption{Screen shots of the app showing a parking lot with two cameras. Status screen shows stalls status in the parking lot.}
    \label{fig:exp}
\end{figure}

The front-end of the system is a smart phone application. It conveys parking information directly to drivers. Screen shots of the application are shown in Figure \ref{fig:exp}.


\section{Experimental Analysis}
In this section the experimental results of the detection module and a comparison against state of the art are presented. First the dataset used in the experiments is introduced, followed by the training method and hyper parameters used. Finally the results and discussion is provided.

\begin{table*}[ht]
\centering
\caption{Results of Single Parking Lot training and testing, with three subsets (PUC, UFPR04, UFPR05)}
\label{table:single}
\resizebox{0.8\textwidth}{!}{%
\begin{tabular}{|c||c|c|c|c|c|c|c|c|c|}
 \hline
  & \multicolumn{3}{|c|}{PUC} & \multicolumn{3}{|c|}{UFPR04} & \multicolumn{3}{|c|}{UFPR05}\\
 \hline
  & AUC & FPR & FNR & AUC & FPR & FNR & AUC & FPR & FNR\\
   \hline
 baseline\_mean & \textbf{0.9998} & 0.004 & 0.0032 & 0.9997 & 0.0044 & 0.0030 & 0.9995 & 0.0078 & \textbf{0.0059}\\
 baseline\_max  & 0.9997 & 0.0053 & 0.0037 & 0.9994 & 0.0050 & 0040 & 0.9991 & 0.0083 & 0.0061 \\
 our\_CNN & 0.9997 & \textbf{0.0007} & \textbf{0.0009} & \textbf{0.9999} & \textbf{0.0001} & \textbf{0.0009} & \textbf{0.9998} & \textbf{0.00008} & 0.0071\\
 \hline
\end{tabular}}
\end{table*}

\subsection{Dataset}
\begin{figure}[ht]
    \centering
    \includegraphics[width=0.2 \textwidth]{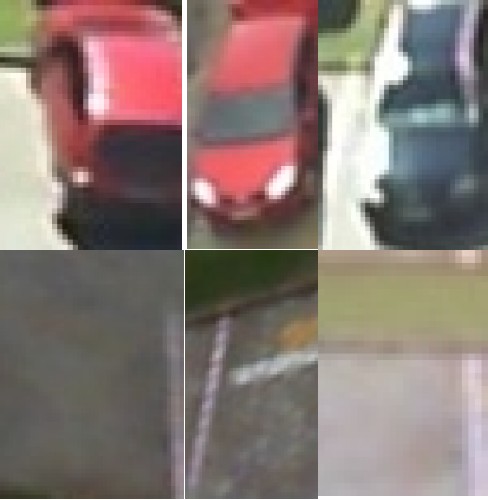}
    \caption{Examples of positive and negative samples extracted from PKLot dataset.}
    \label{fig:dataset}
\end{figure}

The required data for training this network should have a large number of images with cars as positive sample and near equal number of negative samples of images without a car. A diverse dataset is a key factor to have a well generalized network. The dataset should cover cars with different angles and sizes with respect to the camera and in varying weather conditions. \\
The dataset used in the experiments is the PKLot datasets \cite{de2015pklot}. It contains 12,417 images of three parking lots and 695,899 segmented parking spaces in these lots. Two of the parking lots are in the Federal University of Parana (UFPR) and the third is in Pontifical Catholic University of Parana (PUCPR) resulting in three sets of data (UFPR04, UFPR05, PUC). The dataset is general with different appearances for the vehicles, and different conditions such as rainy, sunny, and cloudy. To our knowledge, this is the largest dataset for this task. Figure \ref{fig:dataset} shows some of the images for empty and occupied spaces that were used in the training. Following the same procedure of the authors in \cite{de2015pklot}, each set is split into 50\% training and 50\% testing.

\subsection{Training Method} 
The PKLot dataset is too specific (it only has images of cars and background) to be able to train the whole network from scratch while maintaining generality of the network. Therefore, we initialize the weights in the network from a pre-trained VGGNet-f on ILSVRC-2012 and only fine-tune the network with the data for this task. It helps the network by starting close to the global minimum and reduces the chance of getting stuck in an over-fitted local minimum.\\
Stochastic Gradient Decent was used for fine-tuning with learning rate of 0.01 with learning rate decay, and weight decay of 0.0005. We trained the network for 3000 iterations with mini-batch size of 128. Since the weights of convolutional layers is transferred from a pre-trained network, they should already be suitable for extracting visual features. Therefore, during the training we do not change the weights for first five convolutional layers and limit the training to the top three fully connected layers.

\subsection{Results}
Three sets of experiments are presented, following the same procedure in \cite{de2015pklot} to compare against their baseline classifier:
\begin{itemize}
    \item Single parking lot training and testing, where the network is trained on the training subsets of (PUC, UFPR04, UFPR05) and tested on the corresponding testing subset.
    \item Single parking lot training and multiple parking lot testing, where the network is trained on one of the training subsets and tested on the  testing subsets of the other two parking lots. This experiment ensures that the network is generalized enough to be able to classify parking lots that it has not seen before.
    \item Multiple parking lot training and testing, where the network is trained on all training subsets, and tested on all testing subsets. This experiment provides a measure of how the network is able to cope with variability in the testing subsets.
\end{itemize}

\begin{figure}[ht]
    \centering
    \includegraphics[width=0.45 \textwidth]{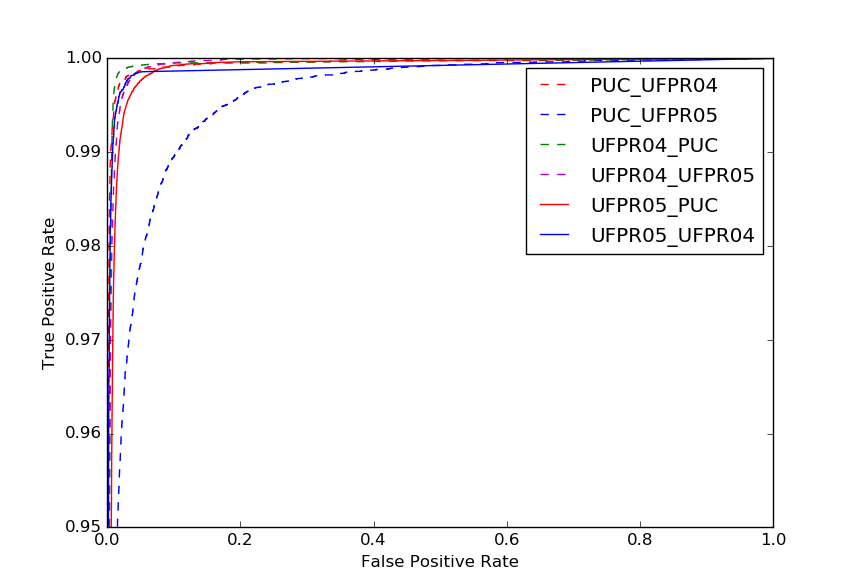}
    \caption{Combined ROC plot of single parking lot training and multiple parking lot testing, format for data is trainLot\_testLot.}
    \label{fig:roc}
\end{figure}

For the quantitative evaluation ROC curves are presented along with False Positive Rate (FPR), False Negative Rate(FNR) and Area Under the Curve(AUC) of ROC curves as evaluation metrics. The experiments are compared to the baseline classifier in \cite{de2015pklot}. Specifically, the classifier with mean rule and max rule as their fusion strategy of multiple SVM classifiers. These classifiers are denoted as baseline\_mean and baseline\_max respectively in the results, while our method is denoted as our\_CNN. Table \ref{table:single} shows the results of the first experiment, and Tables \ref{table:PUC}, \ref{table:UFPR04}, \ref{table:UFPR05} show the results of the second experiment. AUC of our method is between 3 to 5\% better than the previous state of the art. Table \ref{table:all} shows the results of the last experiment. Finally, the ROC curves of different experiments is shown in Figure \ref{fig:roc}.

\begin{table*}[t]
\centering
\caption{Results of Single Parking Lot training on PUC and testing on UFPR04, UFPR05}
\label{table:PUC}
\resizebox{0.6\textwidth}{!}{%
\begin{tabular}{|c||c|c|c|c|c|c|c|c|c|}
 \hline
  & \multicolumn{3}{|c|}{UFPR04} & \multicolumn{3}{|c|}{UFPR05}\\
 \hline
  & AUC & FPR & FNR & AUC & FPR & FNR\\
   \hline
 baseline\_mean & 0.9589 & 0.0427 & 0.1643 & 0.9194 & 0.1574 & 0.1590\\
 baseline\_max  & 0.8826 & 0.0537 & 0.2065 & 0.8363 & 0.2186 & 0.1114 \\
 our\_CNN & \textbf{0.9994} & \textbf{0.009} & \textbf{0.0063} & \textbf{0.995} & \textbf{0.154} & \textbf{0.0061}\\
 \hline
\end{tabular}}
\end{table*}

\begin{table*}[t]
\centering
\caption{Results of Single Parking Lot training on UFPR04 and testing on PUC, UFPR05}
\label{table:UFPR04}
\resizebox{0.6\textwidth}{!}{%
\begin{tabular}{|c||c|c|c|c|c|c|c|c|c|}
 \hline
  & \multicolumn{3}{|c|}{PUC} & \multicolumn{3}{|c|}{UFPR05}\\
 \hline
  & AUC & FPR & FNR & AUC & FPR & FNR\\
   \hline
 baseline\_mean & 0.9713 & 0.2630 & 0.0052 & 0.9582 & 0.1765 & 0.0625\\
 baseline\_max  & 0.9522 & 0.1928 & 0.0467 & 0.9595 & 0.1458 & 0.0708 \\
 our\_CNN & \textbf{0.9989} & \textbf{0.0162} & \textbf{0.0015} & \textbf{0.9992} & \textbf{0.0347} & \textbf{0.0022}\\
 \hline
\end{tabular}}
\end{table*}

\begin{table*}[t]
\centering
\caption{Results of Single Parking Lot training on UFPR05 and testing on PUC, UFPR04}
\label{table:UFPR05}
\resizebox{0.6\textwidth}{!}{%
\begin{tabular}{|c||c|c|c|c|c|c|c|c|c|}
 \hline
  & \multicolumn{3}{|c|}{PUC} & \multicolumn{3}{|c|}{UFPR04}\\
 \hline
  & AUC & FPR & FNR & AUC & FPR & FNR\\
   \hline
 baseline\_mean & 0.9761 & 0.0339 & 0.1826 & 0.9533 & 0.0411 & \textbf{0.2253}\\
 baseline\_max  & 0.9520 & 0.0339 & 0.1826 & 0.9298 & 0.0449 & 0.2920 \\
 our\_CNN & \textbf{0.9982} & \textbf{0.0026} & \textbf{0.1397} & \textbf{0.9989} & \textbf{0} & 0.2886\\
 \hline
\end{tabular}}
\end{table*}

\begin{table}[t]
\centering
\caption{Results of Multiple Parking Lot training and testing [Needs to be filled with correct values]}
\label{table:all}
\resizebox{0.4\textwidth}{!}{%
\begin{tabular}{|c||c|c|c|}
 \hline
  & AUC & FPR & FNR\\
   \hline
 baseline\_mean & 0.9993 & 0.0069 & 0.0072\\
 baseline\_max  & 0.9988 & 0.0339 & 0.0082\\
 our\_CNN & \textbf{0.9997} & \textbf{0.0062} & \textbf{0.00137}\\
 \hline
\end{tabular}}
\end{table}

\begin{figure}[ht]
    \centering
    \begin{subfigure}[b]{0.45\textwidth}
        \centering
        \includegraphics[width=\textwidth]{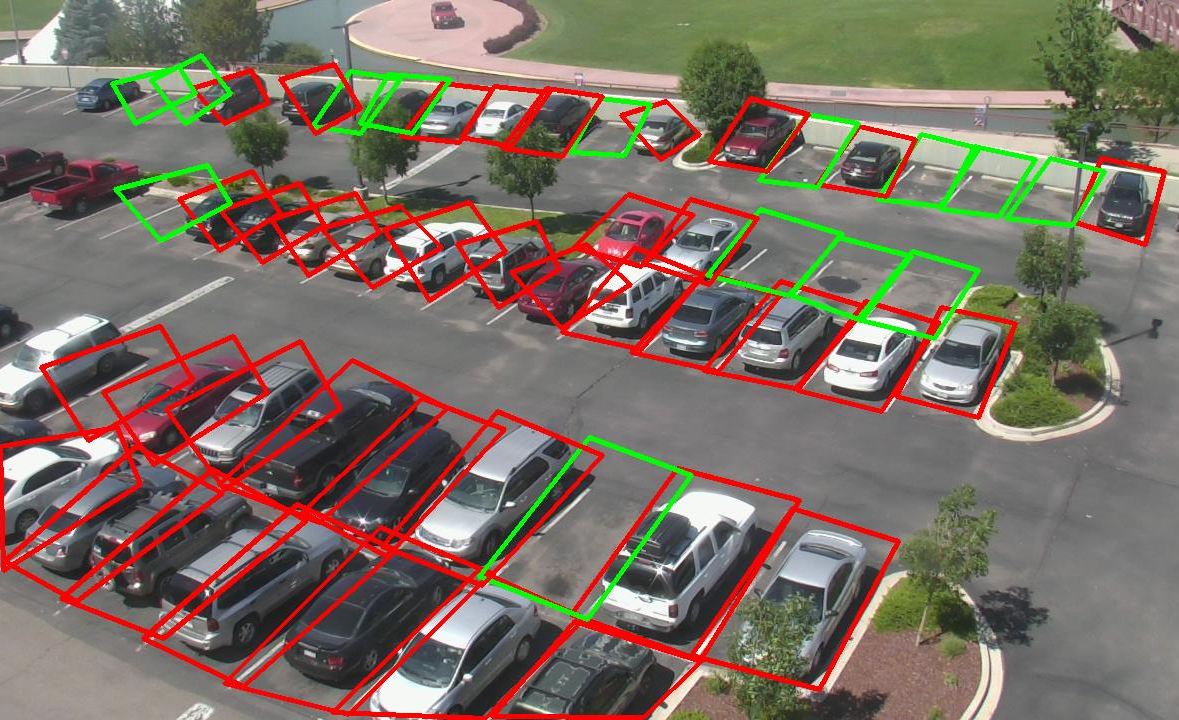}
    \end{subfigure}
    \par \bigskip
    \begin{subfigure}[b]{0.45\textwidth}
        \centering
        \includegraphics[width=\textwidth]{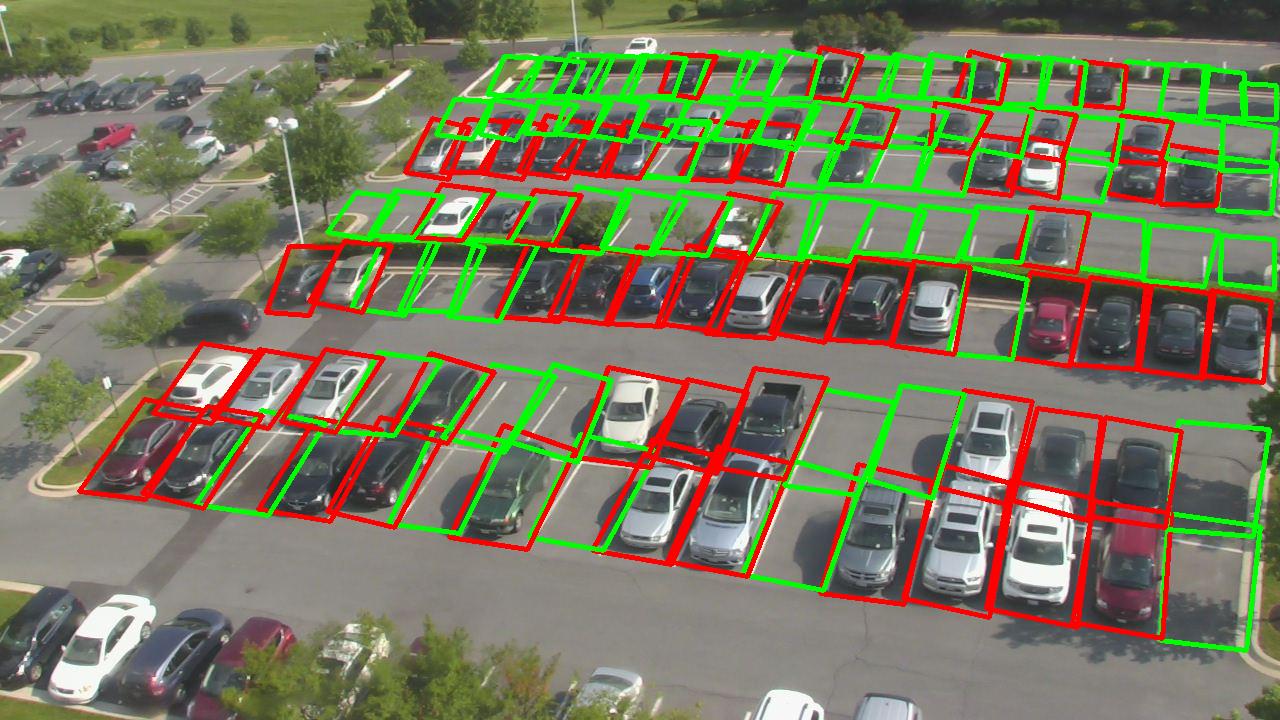}
    \end{subfigure}
    \caption{Qualitative results of the detection system.}
    \label{fig:qual}
\end{figure}
For the qualitative evaluation we have chosen publicly available IP cameras in parking lots that we had no physical control over. This shows our system's robustness to large variety of input data. Sample images of these ip-cameras with labels from our detection system is provided inf Figure \ref{fig:qual}. Figure \ref{fig:exp} is a screen shot of the application for another ip-camera. During our experiments on these cameras, we observed about 5$\%$ missclassification.

%
%
For particular applications, transmitting online footage of cameras over internet may not be desirable due to bandwidth limits or security measures. In these cases, having a local machine that handles the processing is better suited. An immediate question that arises is what is the required computation power for the local machine. Accordingly, we made a set of experiments on different machines and recorded the inference time (the time that our detection module takes to label an image from a stall). Table \ref{table:inftime} summarizes the results. \\
Beside conventional architectures, we also tested the detection module on an embedded architecture. Rapsberry Pi (RPi) devices were chosen as the test platform due to their popularity and low price. We recorded the inference time of $0.22$ on this platform. To put it in a perspective, for a parking lot of 300 stalls, a single RPi can update the status of the whole parking lot in about a minute.   

\begin{table}[ht]
\centering
\caption{Inference time comparison for three different specification. }
\label{table:inftime}
\resizebox{0.49\textwidth}{!}{%
\begin{tabular}{|c|c|}
\hline
	Specification of the Machine & Inference Time (s) \\
    \hline
    GPU Machine (Nvidia GeForce GTX 960) & 3.56e-4 \\ 
    CPU Machine (Intel Core i7-4790K @ 4GHz) & 0.0126 \\ 
    CPU Machine(RPi) (4x ARM Cortex-A53 @ 1.2GHz) & \textbf{0.22} \\ 
    \hline
\end{tabular}}
\end{table}

\section{Conclusion and Future Work}
In this paper we designed and implemented a novel parking management system that uses deep convolutional neural networks for stalls' status detection. We designed the network and trained it on images from the PKLot dataset. We managed to supersede the state of the art performance in this dataset. A complete system for  visual Parking Guidance and Information system including, detection method, server and front-end application is implemented. The application is successfully tested on real parking lot video feeds with no modification made on the pre-existing camera installation.
%
%
%
%
%

In our future work, we want to expand our real world experiments so more conclusive performance results can be obtained. Another objective is to improve the dissemination of the information such that instead of broadcasting raw parking status, the system optimizes the data that each user receives based on their location and parking lot vacancies.

\ifCLASSOPTIONcaptionsoff
  \newpage
\fi

%
\bibliographystyle{plain}
\bibliography{references}

\end{document}